\documentclass[10pt,journal,compsoc]{IEEEtran}

\ifCLASSOPTIONcompsoc
  \usepackage[nocompress]{cite}
\else
  \usepackage{cite}
\fi

\ifCLASSINFOpdf
  \usepackage[pdftex]{graphicx}
\else
  \usepackage[dvips]{graphicx}
\fi

\usepackage{url}

\usepackage{amsmath}

\DeclareMathOperator{\atantwo}{atan2}

\begin{document}
%
\title{IKBT: solving closed-form Inverse Kinematics with Behavior Tree}

\author{Dianmu~Zhang,~\IEEEmembership{Student Member,~IEEE,}
        and~Blake~Hannaford,~\IEEEmembership{Fellow,~IEEE}
\IEEEcompsocitemizethanks{\IEEEcompsocthanksitem D. Zhang and B. Hannaford are with the Department
of Electrical Engineering, University of Washington, Seattle,
WA, 98052.\protect\\
E-mail: blake@uw.edu}}
\IEEEtitleabstractindextext{%
\begin{abstract}
Serial robot arms have complicated kinematic equations which must be solved  to write effective arm planning and control software (the Inverse Kinematics Problem).  Existing 
software packages for inverse kinematics often rely on numerical methods which have 
significant shortcomings. Here we report a new symbolic inverse kinematics solver
which overcomes the limitations of numerical methods, and the shortcomings of previous 
symbolic software packages. We integrate Behavior Trees, an execution planning framework
previously used for controlling intelligent robot behavior, to organize the equation solving
process, and a modular architecture for each solution technique.  
The system successfully solved, generated a LaTex report,
and generated a Python code template for 18 out of 19 example robots of 4-6 DOF.  The system 
is readily extensible, maintainable, and multi-platform with few dependencies. 
The complete package is available with a Modified BSD license on Github.
\end{abstract}

\begin{IEEEkeywords}
AI Reasoning Methods, Behavior Tree, Kinematics
\end{IEEEkeywords}}

\maketitle

\IEEEdisplaynontitleabstractindextext

%
\IEEEpeerreviewmaketitle

\ifCLASSOPTIONcompsoc
\IEEEraisesectionheading{\section{Introduction}\label{sec:introduction}}
\else
\section{Introduction}
\label{sec:introduction}
\fi
%
\IEEEPARstart{S}{ymbolic} inverse kinematics analysis is a non-trivial task critical for operation and design of robot manipulators. 
Considering here serial non-redundant chains of up to six degrees of freedom, the inverse kinematics computation
takes the desired end-effector pose as input (typically as a homogeneous transform), and solves for joint angles or joint displacements from the forward
kinematic equations. 

While there are many existing packages for numerical inverse kinematics \cite{matlab} \cite{ROB:ROB4620070406}, these share common limitations
including, finding only one of the multiple solutions, requirement of a starting value, dependence on the starting value, 
and problems with convergence near singular configurations.

Several groups have attempted to automate symbolic inverse kinematics analysis starting in the  1990's \cite{Herrera-Bendezu} \cite{Halperin}, which laid the foundation for our work.  Their work is reviewed and compared with ours in detail in the Discussion section. 

In this work we develop an automated symbolic inverse kinematics package with the following goals:
\begin{itemize}

\item Create a highly extensible and modifiable architecture using a flexible behavior tree for 
    solution logic and a modular design. 
    \item Provide convenience features such as automatic documentation and code generation. 
	\item Implementation in a modern open-source, cross-platform, programming language (Python)
    \item Require minimal dependencies outside of the standard Python distribution (mainly the symbolic
    manipulation package {\tt sympy}).

\end{itemize}

Unlike others' work that mostly used a linear and inflexible method, we adapt Behavior Tree - popular in video game AI - to construct an expert system, ``IKBT", that has the logical reasoning power to solve inverse kinematics symbolically without human supervision. A Behavior Tree uses a  directed graph to model intelligent agent behavior \cite{lim2010evolving, marzinotto2014towards, IROS17Colledanchise, ISR16Colledanchise}. Behavior Trees have the advantages of composability and scalability compared  to finite state machines.

The main contributions  of this work are:

\begin{itemize}

\item We compactly encode the inverse kinematics logic and strategy in a Behavior Tree (see Work Flow and Architecture section).

\item We code each knowledge-based solver into a modular leaf, forming a ``tool box" which is organized by the Behavior Tree (see  Transformations and Solvers section).

\item IKBT generates a dependency {\it graph} of joint variables in the solutions, includes all possible poses. Tracking these dependencies facilitates grouping variables into distinct solutions, essential to downstream control softwares for robots (see Solution Graph section).

\item IKBT successfully solves complicated robots, such as the 6-DOF commercial robot manipulator PUMA 560 and successfully solved 18 out of 19 test robots (95\% success rate) (see Results section).

\item On average, IKBT generates symbolic solutions and source code in a few minutes on a normal PC. The same work load often takes a human expert hours to complete.

\item IKBT generates a  report of its results in \LaTeX, and generates code in Python and C++ creating functions to implement the derived solutions with domain (reachability) checking of numerical inputs(see Pose Validation section). 

\item Inverse kinematics solutions from IKBT are verifiable with numerical values(see Result Verification subsection under Results).  
\end{itemize}



\section{Work Flow and Architecture}
\subsection{Work Flow}

\begin{figure}
	\centering
	\includegraphics[width = 0.9\linewidth]{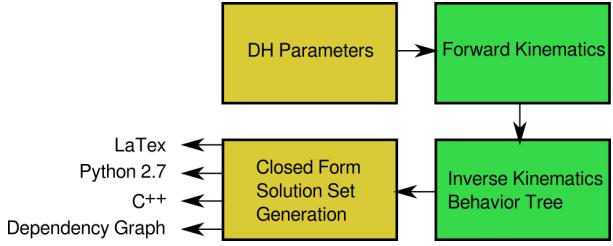}
    \caption{\textbf{Work Flow.} Forward kinematics module computes symbolic kinematic equations to be solved ($T_d = T_s$) given the input DH parameters. Then the equations are evaluated for closed-form inverse kinematics solutions to each joint variable. Upon solving a robot, along with the solutions, a dependency graph, Latex report, and Python/C++ code are generated as convenience feature.}\label{fig-flow}
\end{figure}

As shown in Fig. \ref{fig-flow}, the system input takes symbolic Denavit--–Hartenberg (DH) parameters and 
calculates symbolic forward kinematic equations in the form of a 4x4 homogeneous transformation. 
For a 6-DOF robot, the transformation matrix $T^6_0$ is computed as

\begin{equation} \label{eq1}
T_s  =  T^6_0 = T^1_0 T^2_1 T^3_2 T^4_3 T^5_4 T^6_5
\end{equation}

By convention, each transformation matrix takes a coordinate from the subscript frame and transforms it to the superscript frame 
($T^6_0$ transforms from frame 0 to frame 6). 

We denote the desired robot end effector pose as $T_d$.  Then the inverse kinematics problem can be stated as solving 
\begin{equation} \label{eq2}
T_d = T^6_0(q_1, \dots, q_6)
\end{equation}
(where $q_i$ are the unknown joint variables) 
for all sets of joint variables ($q_i = \theta_i$ or $d_i$) which satisfy \ref{eq2}.   
Related equations which can be used to find soluble equations include:

\begin{equation} \label{eq3}
[T^1_0]^{-1}T_d = [T^1_0]^{-1}T_s
\end{equation}

\begin{equation} \label{eq4}
[T^2_1]^{-1}[T^1_0]^{-1}T_d = [T^2_1]^{-1}[T^1_0]^{-1}T_s
\end{equation}

\begin{equation} \label{eq5}
[T^{n-1}_{n-2}]^{-1} ... [T^1_0]^{-1} T_d =[T^{5}_{4}]^{-1} ... [T^1_0]^{-1} T_s
\end{equation}

IKBT first symbolically calculates and simplifies these intermediate results
(\ref{eq3} - \ref{eq5}) to augment (\ref{eq1}).  
Each of these matrix equations creates 12 scalar equations (one for each element in the first three rows) which can be searched for solvable equations.  
After generating these matrices, 
each scalar equation is categorized by the number of unsolved joint variables, $q_i$, 
into three lists according to the number of unsolved variables in each equation (1, 2, and 3-or-more unknowns). As each variable is solved, this scan is repeated.  The current toolbox examines equations containing 1- and 2-unsolved variables. 
   
The Behavior Tree's leaf nodes transform, or  identify and solve, a particular kind of equation (See detail in the following section "Transforms and Solvers").  For example, one pair of nodes identifies and solves scalar equations of one-unknown of the form $A = \sin(B\theta_j+C)$ or $A = \cos(B\theta_j+C)$
where $A,B,C$ are known expressions, and $\theta_j$ is unknown. 

After all joint variables are solved, the solution graph and the solution vectors ($2^n$ joint vectors in symbolic form correctly associating the multiplicity of each variable) are constructed. A  \LaTeX report, C++ and Python code, containing symbolic solutions for all possible poses, is also generated.  

\subsection{Architecture}
Behavior Trees have been explored in the context of humanoid robot control \cite{marzinotto2014towards, colledanchise2014performance, bagnell2012integrated},
collaborative robotics \cite{guerin2015framework,ISR16Colledanchise}, 
and as a modeling language for intelligent robotic surgical procedures \cite{DYHu-BT,DBLP:journals/corr/HannafordHZL16}.  

\begin{figure*}[ht]
	\centering
	\includegraphics[width = 0.9\textwidth]{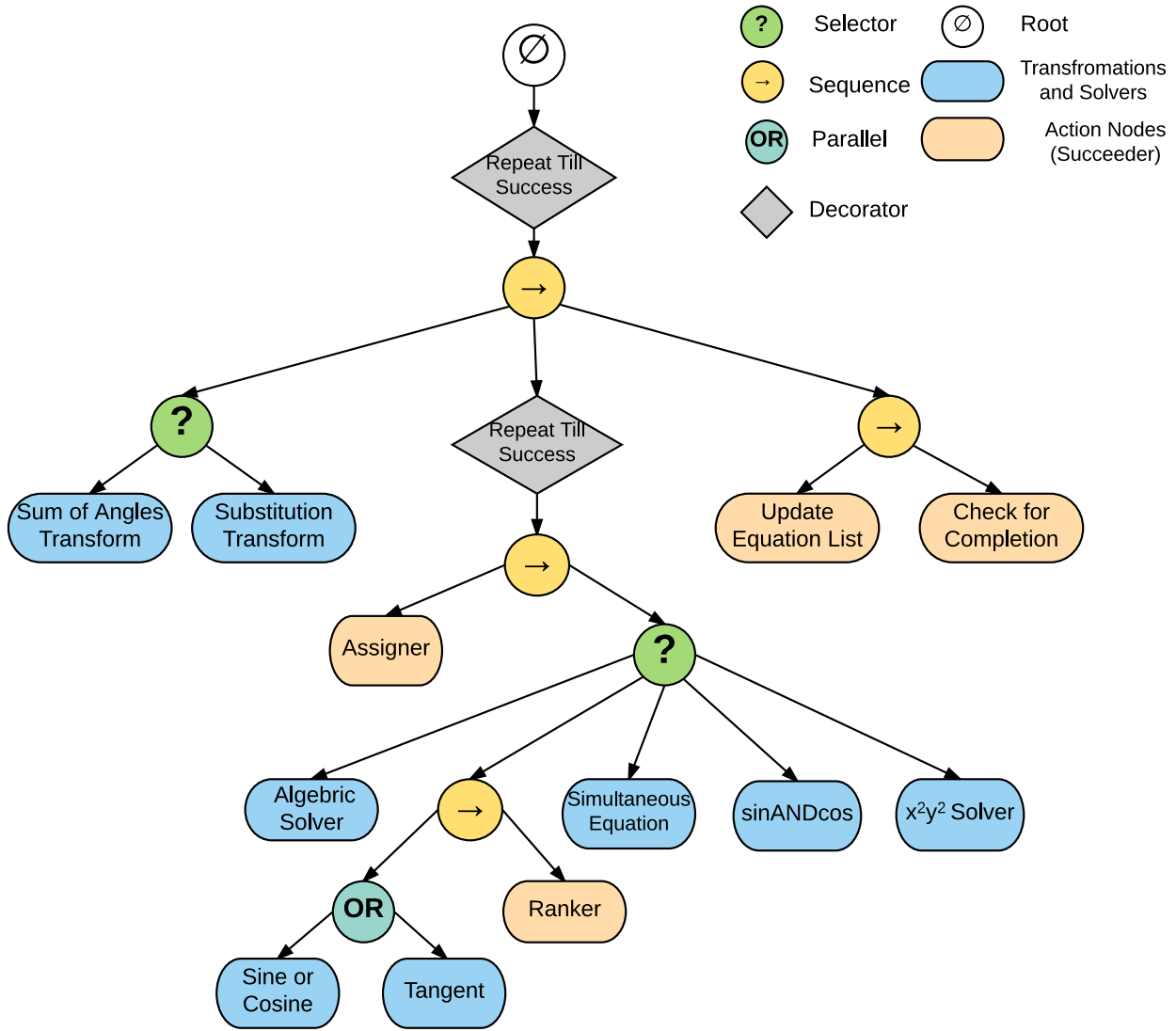}
    \caption{\textbf{IKBT Structure.} Node type explanation: Action nodes (leaves) carry out specific tasks, and returns SUCCESS or FAILURE.Succeeder is a special type of action nodes that only returns SUCCESS. Selector node ticks its children in turn, returns SUCCESS and stops if one of the children succeeds, otherwise returns FAILURE. Sequence node only returns SUCCESS if all its children succeed. Parallel node tries out all its children regardless of their return status, returns SUCCESS if any child succeeds.}
    \label{fig-BTStructure}
\end{figure*}

The work reported here is the first to our knowledge to use Behavior Trees to encode algorithms for reasoning about and solving mathematical equations symbolically.
When implementing intelligent behavior with Behavior Trees, 
the designer of a robotic control system breaks the task down into modules (Behavior Tree leaves) 
which return either ``success" or ``failure" when called by parent nodes.  Higher level nodes 
define composition rules to combine the leaves including: Sequence, Selector, and Parallel node types which 
also return ``success'' or ``failure''.
A Sequence node defines the order of execution of leaves and returns success if all leaves succeed in order. A Selector node (called ``Priority" by some authors) tries leaf behaviors in a fixed order, 
returns success  when a node succeeds,  and returns failure if all leaves fail. We also implemented a ``Parallel'' node (represented as ``OR" in Fig. \ref{fig-BTStructure}), which executes all leaves regardless of their return status, and returns success if any one of the leaves succeeds. The IKBT structure used for our current results is shown in Fig.\ref{fig-BTStructure}.

Before solving, IKBT looks through 1- and 2- unknown equation lists, and applies sum-of-angle and 
substitution transformations which may reduce number of unknown variables. 
The ``Assigner" node assigns the current variable to all solvers. 
For each joint variable, it tries out all solvers in the toolbox until it is solved 
(or we reach the maximum trial number). 
When a joint variable is solved, the solver marks it as solved, and reduces the number of unsolved variables by one, for all equations involved this variable.  When multiple solvers can solve a joint variable, the solutions are compared using a ``Ranker" node which selects preferred solution forms over others.  For example, $\theta_4=$atan2($y,x$) is preferred over
$\theta_4 = \mathrm{arcsin}(y/r)$ because it has only one solution. 
IKBT repeats this until all variables are solved. Finally the solutions, report, code generation, 
and dependency graph are generated. 

\section{Transformations and Solvers} 
In the following subsections, $\theta_i$ and $d_i$ represent rotatory and prismatic joint variables ($q_i$).
$a$, $b$, $c$, etc., stand for known constant DH parameters.

\subsection{Transformations}
Transformation nodes make equations easier to solve by reducing the number of unknown variables.

\begin{enumerate}
\item Sum of angle transform
\[
\sin(\theta_x \pm \theta_y) \rightarrow \sin(\theta_{xy})
\]
\[
\cos(\theta_x \pm \theta_y) \rightarrow \cos(\theta_{xy})
\]
Although the sum-of-angle simplification is done by sympy's {\tt simplify} operation, creation of a new variable ($\theta_{xy}$) is done by this node. 

\item Substitution transform: looks for two equations such
that one contains the other, and replaces the partial expression with a unknown value. For example, the following pair of equations:
\[
\sin(\theta_x)  + a \cdot \cos(\theta_y) = b
\]

\[
a \cdot \cos(\theta_y) = c
\]

The first equation is transformed to:

\[
\sin(\theta_x) + c = b
\]

Eliminating one unknowns so that $\theta_x$ can be solved.
\end{enumerate}

\subsection{Rule-based Solvers}

The IKBT  contains a set of solvers that identifies an expression that fits a rule set and return the respective solutions. These rules are used by human experts when solving inverse kinematics problems, and not are specific to any DOF or robot configuration. 

\begin{enumerate}
\item algebraic solver 
\\Identifies pattern 
\[
a + b \theta = c
\]
where $b \neq 0$.  Solves for 
\[
\theta = \frac{c - a}{b} 
\]

as well as 

\[
a + b d_x = c
\]

giving
\[
d_x = \frac{c - a}{b} 
\]

\item sine or cosine solver 
\\Identifies pattern 
\[
\sin(\theta) = a, \quad \quad \cos(\theta) = b
\]
Solves for 
\[
\theta = \arcsin(a), \quad \quad \theta = \arccos(b) 
\]
or 
\[
\theta = \pi - \arcsin(a), \quad \quad \theta = - \arccos(b)
\]

\item tangent solver
\\identifies a pattern in two equations containing
\[
\sin(\theta) = aC_1 \quad and \quad \cos(\theta) = bC_2
\]

If neither $C_1$ or $C_2$ contain unsolved variables:
\[
\theta = \atantwo (\frac{a}{C_1}, \frac{b}{C_2}) 
\]

Sometimes $C_1$ and $C_2$ contain common unsolved variables, which can be canceled out by division. In this case we use a new coefficient $C$:

\[
C = \frac{C_1}{C_2}
\]

\[
\theta = \atantwo (\frac{a}{C}, b) \quad C > 0
\]

\[
\theta = \atantwo (-\frac{a}{C}, -b) \quad C < 0
\]

Terms which are solvable by tangent solver are often also solvable by sine or cosine solver. As shown in Fig \ref{fig-BTStructure}, IKBT takes this into consideration by comparing the solutions from the above-mentioned solvers, and determines the optimal solution. This selection is done by the Ranking node. 

\item Sine and cosine solver

Identifies 
\[
a\cdot \sin(\theta) + b \cdot \cos(\theta) = 0 
\]
giving 
\[
\theta = \atantwo (-b, a) , \quad \quad \theta = \atantwo (-b, a) + \pi
\]

as well as 
\[
a\cdot \sin(\theta) + b \cdot \cos(\theta) = c
\]
giving
\[
\theta = \atantwo (a,b) + \atantwo (\pm \sqrt{a^2 + b^2 - c^2}, c)
\]

\item Simultaneous equation solver \cite{Herrera-Bendezu}
Identifies two equations:

\[ a \sin(\theta) + b \cos(\theta) = c  \qquad
a \cos(\theta) - b \sin(\theta) = d   \]

Giving the solution: \[ \theta = \atantwo (ac - bd, ad + bc)\]

\item $x^2 y^2$ solver 

Identifies two equations that contain $P_x$, $P_y$, and/or $P_z$, that can be squared and added together to cancel out unsolved variables (other than the intended  variable), and get a new equation with pattern \cite{Craig:1989:IRM:534661} :

\[ -\sin(\theta) P_x + \cos(\theta) P_y = d  \] 

Giving solutions: 
\[ 
\theta = \atantwo (P_y, P_x) - \atantwo (d, \pm \sqrt{{P_x}^2 + {P_y}^2 - d^2})
\]

\end{enumerate}



\section{Solution Graph}

\subsection{Origins of Dependency}
The solutions produced by inverse kinematics are typically interdependent in that results obtained early in the process are used to solve later results. For example, one may have 
\[
\theta_{4} = \operatorname{asin}{\left (\frac{1}{l_{4}} \left(Pz - l_{3} + l_{5} \cos{\left (\theta_{45} \right )}\right) \right )}
\]
in which $\theta_4$ depends on $\cos(\theta_{45})=\cos(\theta_4+\theta_5)$ as well as some constants.  In principle, it is possible to substitute these dependencies until there are no joint variables on the right hand side, 
but this makes the solutions difficult to compare with previously published hand solutions.  
In the above example, there are two solutions to the $\operatorname{asin}()$ operator, and additional multiplicity could come from the solution method for
$\theta_{45}$.

Thus the  two sources of multiple solutions are: A) each joint variable may have multiple solutions due to its solver's characteristic; 
and B) dependence of the solution on other solved joint variables.  For example, 
\begin{equation} \label{eg1}
\cos(\theta_1) = a  
\end{equation} 
\[
\theta_1 = \left \{
	\begin{array}{cr}
    \theta_{1s1} =& \operatorname{acos}(a)\\
    \theta_{1s2} = & -\operatorname{acos}(a)
    \end{array} \right .
\]
(where we have used the subscript $s$ to separate joint numbers from solution numbers.  For example, $\theta_{1s2}$ means solution 2 of $\theta_1$).
Now that $\theta_1$ is solved, we can write
\begin{equation} \label{eg2}
\sin (\theta_2) + \sin (\theta_1) = b
\end{equation} 
\[
\theta_2 = \left \{
	\begin{array}{rr}
     \theta_{2s1} =& \arcsin(b - \sin(\theta_{1s1}))\\
     \theta_{2s2} =& \pi - \arcsin(b - \sin(\theta_{1s1}))\\
     \theta_{2s3} =& \arcsin(b - \sin(\theta_{1s2}))\\
     \theta_{2s4} =& \pi - \arcsin(b - \sin(\theta_{1s2}))
    \end{array} \right .
\]

In the resulting graph (shown in Fig. \ref{fig-sample-graph} a), each joint solution (e.g. $\theta_{2s1}$, $\theta_{2s2}$, etc.) is a node. A parent node is the node that appears in another node's solution expressions, in this example, $\theta_{1s1}$ is the parent of $\theta_{2s2}$ . A node and its parent/child node are connected with an edge.

\subsection{Redundancy Detection and Dependency Tracking}

When building a dependency graph, we implemented redundancy elimination to ensure the correct relations between joint variables. Redundancy is defined as a dependency that traced back to a higher level parent can be mediated by a lower level and direct parent. If a joint variable $\theta_6$ has the following solution:
\[
\theta_6 = atan2(a + \cos(th_4), \sin(th_5) b) 
\]
And $\theta_5$ has solution:
\[
\theta_5 = \arccos(l + \cos(\theta_4))
\]

Though the solution of $\theta_6$ involves both $\theta_4$ and $\theta_5$, its dependency to $\theta_4$ is redundant. Given that $\theta_5$ is also dependent on $\theta_4$, the effects of choosing different $\theta_4$ values (if applicable) on $\theta_6$ are conveyed through $\theta_5$. Therefore, when building a graph, only the edges between direct child-parents are added, in this case, Edge $(\theta_6, \theta_5)$ and Edge$(\theta_5, \theta_4)$. Shown in Fig. \ref{fig-sample-graph} b).

Classical search algorithms (breadth-first search and depth-first search) are used to traverse the graph and find correct ancestor nodes, where the current variable is the start point and the ancestors are the goals.

\subsection{Grouping Variables}
As required by many planning and control softwares, IKBT is capable of grouping variables into solution sets that have all possible joint configurations for the given end-effector configuration. To generate correct sets of solutions, the following steps are carried out to match the variables: First, all parents nodes are extracted from each solution expression, and formed subsets of variables. Secondly, the subsets are sorted by size of their content. Search starts from the largest subsets, and looks for the variables that are a part of the joint space, but not in the set, till all variables are found. A scoring system is applied on all subsets (other than the starting set) to focus the search on the more likely candidate first.   

Using the Fig. \ref{fig-sample-graph} a) as an example, the solutions can be grouped into: [$\theta_{1s1}$, $\theta_{2s1}$], [$\theta_{1s1}$, $\theta_{2s2}$], [$\theta_{1s2}$, $\theta_{2s3}$], and [$\theta_{1s2}$, $\theta_{2s4}$].

\subsection{Graph Representation}

The multiple dependencies can be linked by a common dependency further up, or they can be independent. Although traditionally this structure is represented as a tree, we discovered cases in which variables have multiple independent ``parents'' and thus a graph is required instead. 

For example, $\theta_1$ and $\theta_2$ are independent to each other:

\begin{align*}
\theta_{1s1} =& \arcsin(a)\\
\theta_{1s2} =& -\arcsin(a) + \pi \\
\theta_{2s1} =& \arccos(b) \\
\theta_{2s2} =& -\arccos(b) 
\end{align*}

And $\theta_3$ depends on both $\theta_1$ and $\theta_2$:

\begin{align*}
\theta_{3s1} =& \arccos(a + cos(\theta_{1s1}) + atan2(b, sin(\theta_{2s1}) c) \\
\theta_{3s2} =& \arccos(a + cos(\theta_{1s1}) + atan2(b, sin(\theta_{2s2}) c) \\
\theta_{3s3} =& \arccos(a + cos(\theta_{1s2}) + atan2(b, sin(\theta_{2s1}) c) \\
\theta_{3s4} =& \arccos(a + cos(\theta_{1s2}) + atan2(b, sin(\theta_{2s2}) c) \\
\end{align*}

\begin{figure}
    \centering
    \includegraphics[width = 0.8\linewidth]{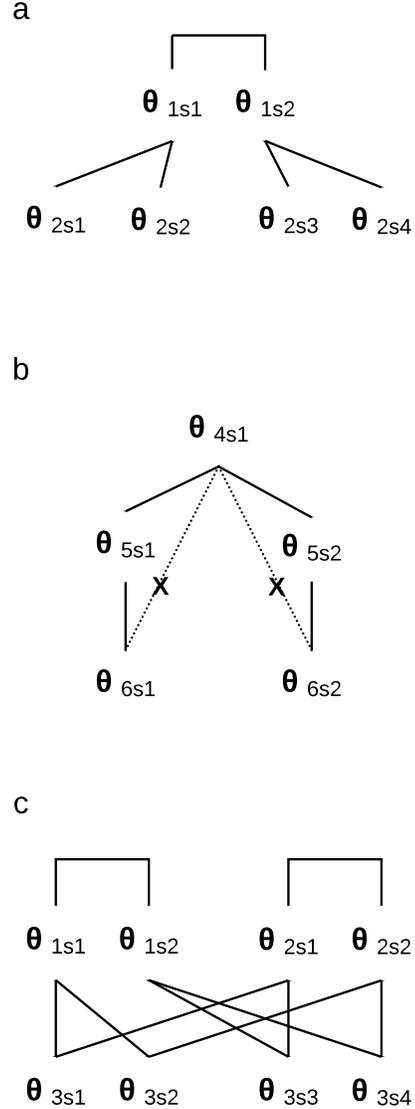}
    \caption{\textbf{Sample Solution Graph.} a) Simple example solution graph explains the origins of multiplicity. b) Redundancy pruning, 'x' marks the dependent relations that are not included in the graph.  c) Example of a case of variables with multiple independent parents.}
    \label{fig-sample-graph}
\end{figure}

The dependency graph is shown in Fig. \ref{fig-sample-graph} c).

One example robot solution in Results section shows the necessity of using graph representation. 
\section{Verification}
\subsection{Solution Verification} \label{solution-verify}
\begin{figure*}[ht]
	\centering
    \includegraphics[width = 1\textwidth]{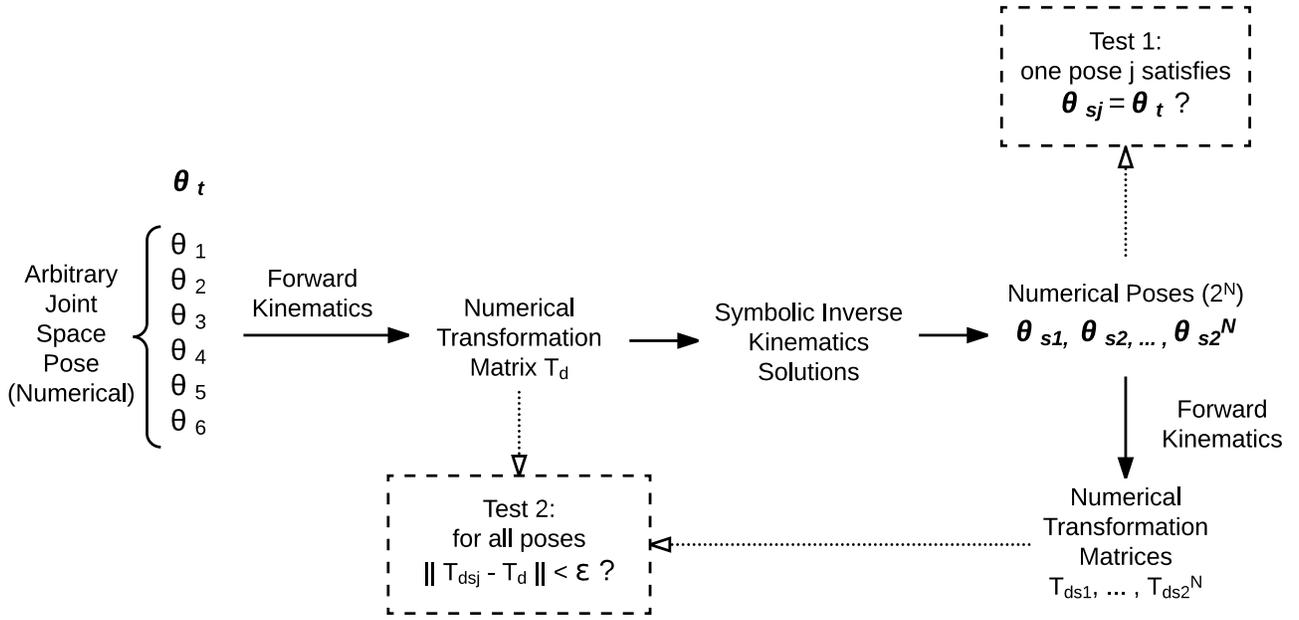}
    \caption{\textbf{Result Verification.} A numerical 4x4 homogeneous transformation matrix, $T_d$, is constructed from a reachable pose. Numerical joint space poses are computed from $T_d$ using the closed-form solutions.  For each solution pose, forward kinematics is calculated. The resulting transform matrices are compared against the original matrix, matching value is indicative of correct IKBT inverse kinematics analysis.} 
    \label{fig-verify}
\end{figure*}

To prove that the inverse kinematics solution equations from IKBT are correct, we conducted the following verification process, as shown in Fig. \ref{fig-verify} . First, we constructed a valid numerical transformation matrix from a reachable pose. Then used numbers from the transformation matrix and the inverse kinematics solutions equations to get the numerical values for each pose. If the inverse kinematics solutions are correct, one of the poses should match the starting pose value. Next, compute the forward kinematics using each numerical pose. If the resulted transformation matrix is the same (within a stringent range) as the starting matrix, then we can safely draw the conclusion that this pose has correct inverse kinematics solution.

\subsection{Pose Validation}
If a pose is not reachable by the 
robot (for example due to distance of a point extending beyond the length of the arm, but not considering joint limits), at least in generated code output, the solution must 
have a means to detect this case.  
In inverse kinematic solution equations, unreachable poses generate intermediate 
values outside the domain of transcendental functions, for example:
\[
\theta_2 = \arcsin(x) \qquad x = 1.2
\]
or would require complex joint angles:
\[
d_3 = \sqrt{x} \qquad  x = -5
\]

Both the C++ and Python output modules of IKBT generate code which checks numerical arguments of inverse trig functions and square roots for such cases and returns a flag to indicate an unreachable pose.

\section{Results}
\subsection{General performance}
We tested IKBT on many sets of DH parameters, representing serial arm robot designs (including commercial robots, and solved design examples from student homework), the successful solving rate is listed in Table \ref{resultstable}. As the DOF number increases, the problem becomes more complex and the success rate decreases. In general it solves most of the robots, up to 6 DOF. Note that IKBT can solve robots regardless of their configurations, e.g. IKBT does not require robots having three intersecting axes.

\begin{table}[b]
\label{results}
\begin{center}
\begin{tabular}{|c|c|c|}
\hline
number of DOF & Test examples & Solved \\
\hline
4 & 4 & 4 \\
\hline
5 & 10 & 10\\
\hline
6 & 5 & 4\\
\hline
\end{tabular}
\end{center}
\caption{IKBT test results}\label{resultstable}
\end{table}

Source code can be found at: \url{https://github.com/uw-biorobotics/IKBT}. The DH parameters of all these robots are stored as part of the source code (in {\tt ik\_robots.py}), for purpose of testing and reproducing the results. Instructions are on the GitHub page. 

\subsection{Sample solutions PUMA 560}

\begin{table}[h]
\caption{PUMA560 DH parameters}
\label{puma}
\begin{center}
\begin{tabular}{|c|c|c|c|c|}
\hline
Link & $\alpha_{N-1}$ & $a_{N-1}$ & $d_N$ & $\theta_N$ \\
\hline
1 & 0 & 0 & 0 & $\theta_1$\\
\hline
2 & $-\pi/2$ & 0 & 0 & $\theta_2$\\
\hline
3 & 0 & $a_2$ & $d_3$ & $\theta_3$\\
\hline
4 & $-\pi/2$ & $a_3$ & $d_4$ & $\theta_4$\\
\hline
5 & $\pi/2$ & 0 & 0 & $\theta_5$\\
\hline
6 & $-\pi/2$ & 0 & 0 & $\theta_6$\\
\hline
\end{tabular}
\end{center}
\end{table}

Here PUMA560 is used as an example to illustrate how IKBT solves inverse kinematics problems. PUMA 560 was a commercial robot with six rotary joints and four joint offsets, well-known for its challenging inverse kinematics properties.  The PUMA 560 has three axes intersecting at its wrist. The known variables are: $a_2$, $a_3$, $d_3$, and $d_4$.

First forward kinematics was calculated from DH parameters:

$$
T_d = T_s  
$$
$$
T_0^6 = T_0^1 T_1^2 T_2^3 T_3^4 T_4^5 T_5^6
$$
(where $T_d$ is the ``desired position", $T_s$ symbolic expressions, $T_0^6$ transformation matrix from frame 0 to 6)
\[
T_0^6 = 
\begin{bmatrix}
r_{11} & r_{12} & r_{13} & P_x \\
r_{21} & r_{22} & r_{23} & P_y \\
r_{31} & r_{32} & r_{33} & P_z \\
0 & 0 & 0 & 1
\end{bmatrix}
= 
\begin{bmatrix}
v1 & v2 & v3 & v4
\end{bmatrix}
\]

$v_1 =$
\[
\begin{bmatrix}
c_{6} (- c_{1} s_{23} s_{5} + c_{5} (c_{1} c_{23} c_{4} + s_{1} s_{4})) - s_{6} (c_{1} c_{23} s_{4} - c_{4} s_{1}) \\
c_{6} (c_{5} (- c_{1} s_{4} + c_{23} c_{4} s_{1}) - s_{1} s_{23} s_{5}) - s_{6} (c_{1} c_{4} + c_{23} s_{1} s_{4}) \\
- c_{6} (c_{23} s_{5} + c_{4} c_{5} s_{23}) + s_{23} s_{4} s_{6} \\
0
\end{bmatrix}
\]

$v_2 = $
\[
\begin{bmatrix}
- c_{6} (c_{1} c_{23} s_{4} - c_{4} s_{1}) - s_{6} (- c_{1} s_{23} s_{5} + c_{5} (c_{1} c_{23} c_{4} + s_{1} s_{4})) \\
- c_{6} (c_{1} c_{4} + c_{23} s_{1} s_{4}) - s_{6} (c_{5} (- c_{1} s_{4} + c_{23} c_{4} s_{1}) - s_{1} s_{23} s_{5})\\
c_{6} s_{23} s_{4} + s_{6} (c_{23} s_{5} + c_{4} c_{5} s_{23}) \\
0
\end{bmatrix}
\]

\[
v_3 = \begin{bmatrix}
- c_{1} c_{5} s_{23} - s_{5} (c_{1} c_{23} c_{4} + s_{1} s_{4}) \\
- c_{5} s_{1} s_{23} - s_{5} (- c_{1} s_{4} + c_{23} c_{4} s_{1})\\
- c_{23} c_{5} + c_{4} s_{23} s_{5}\\
0
\end{bmatrix}
\]

\[
v_4 = \begin{bmatrix}
a_{2} c_{1} c_{2} + a_{3} c_{1} c_{23} - c_{1} d_{4} s_{23} - l_{3} s_{1}\\
a_{2} c_{2} s_{1} + a_{3} c_{23} s_{1} + c_{1} l_{3} - d_{4} s_{1} s_{23}\\
- a_{2} s_{2} - a_{3} s_{23} - c_{23} d_{4} \\
1
\end{bmatrix}
\]

where $c_1 = cos(\theta_1)$, $s_{23} = sin(\theta_2 + \theta_3)$ etc.

\begin{figure*}[ht]
    \centering
    \includegraphics[width = 0.7\textwidth]{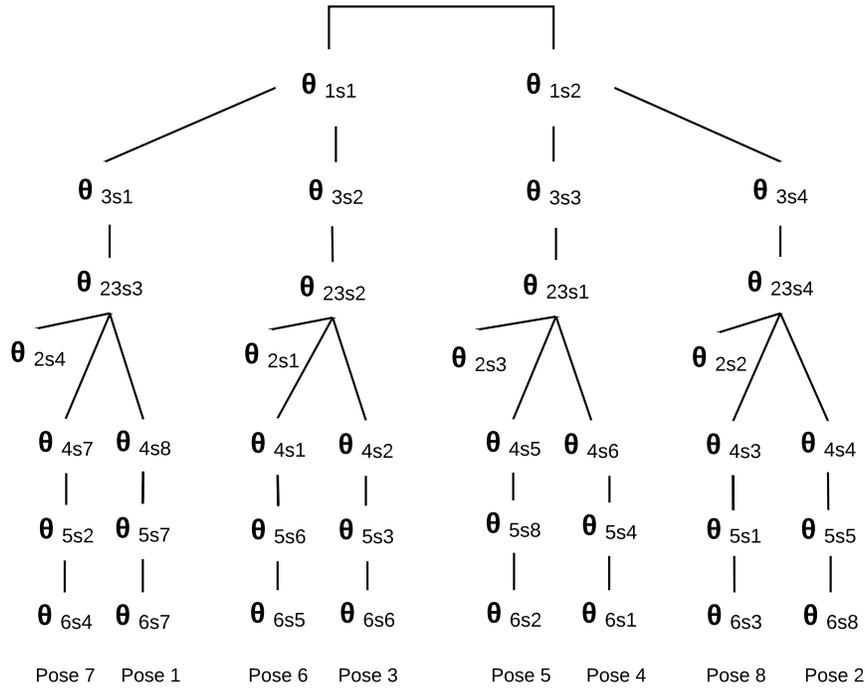}
    \caption{\textbf{PUMA 560 Solution Graph.} }
    \label{fig-Puma-sol}
\end{figure*}

IKBT solved these variables in the following order:
\begin{enumerate}
\item $\theta_1$, chosen solver: sinANDcos
\begin{align*}
\theta_{1s1} = & \operatorname{atan2}{ (Px,- Py  )} + \\
& \operatorname{atan2}{ (\sqrt{Px^{2} + Py^{2} - d_{3}^{2}},- d_{3}  )} \\
\theta_{1s2} = & \operatorname{atan2}{ (Px,- Py  )} + \\
& \operatorname{atan2}{ (- \sqrt{Px^{2} + Py^{2} - d_{3}^{2}},- d_{3}  )} 
\end{align*}

\item $\theta_3$, chosen solver: $x^2y^2$
\begin{align*}
\theta_{3s1} &= \operatorname{atan2}{ (- 2 a_{2} d_{4},2 a_{2} a_{3}  )} + \\
& \operatorname{atan2} (\sqrt{s - (t+ (Px \cos{ (\theta_{1s1}  )} + Py \sin{ (\theta_{1s1}  )})^{2})^{2}}, \\
& t+ (Px \cos{ (\theta_{1s1}  )} + Py \sin{ (\theta_{1s1}  )})^{2}  ) \\
\theta_{3s2} &= \operatorname{atan2}{ (- 2 a_{2} d_{4},2 a_{2} a_{3}  )} + \\
& \operatorname{atan2} (- \sqrt{s - (t+ (Px \cos{ (\theta_{1s1}  )} + Py \sin{ (\theta_{1s1}  )})^{2})^{2}}, \\
& t+ (Px \cos{ (\theta_{1s1}  )} + Py \sin{ (\theta_{1s1}  )})^{2}  ) \\
\theta_{3s3} &= \operatorname{atan2}{ (- 2 a_{2} d_{4},2 a_{2} a_{3}  )} + \\
& \operatorname{atan2} (\sqrt{s - (t+ (Px \cos{ (\theta_{1s2}  )} + Py \sin{ (\theta_{1s2}  )})^{2})^{2}}, \\
& t+ (Px \cos{ (\theta_{1s2}  )} + Py \sin{ (\theta_{1s2}  )})^{2}  ) \\
\theta_{3s4} &= \operatorname{atan2}{ (- 2 a_{2} d_{4},2 a_{2} a_{3}  )} + \\
& \operatorname{atan2} (- \sqrt{s - (t+ (Px \cos{ (\theta_{1s2}  )} + Py \sin{ (\theta_{1s2}  )})^{2})^{2}}, \\
& t+ (Px \cos{ (\theta_{1s2}  )} + Py \sin{ (\theta_{1s2}  )})^{2}  ) 
\end{align*}

where,
\begin{align*}
s = & 4 a_{2}^{2} a_{3}^{2} + 4 a_{2}^{2} d_{4}^{2} \\
t = & Pz^{2} - a_{2}^{2} - a_{3}^{2} - d_{4}^{2} 
\end{align*}

\item $\theta_{23}$, chosen solver: simultaneous equation
\begin{align*}
\theta_{23s1}& = \operatorname{atan2} (Pz (- a_{2} \cos{ (\theta_{3s3}  )} - a_{3}) - \\
& (- Px \cos{ (\theta_{1s2}  )} - \\
& Py \sin{ (\theta_{1s2}  )}) (a_{2} \sin{ (\theta_{3s3}  )} - d_{4}), \\
& Pz (a_{2} \sin{ (\theta_{3s3}  )} - d_{4}) + \\
& (- Px \cos{ (\theta_{1s2}  )} - Py \sin{ (\theta_{1s2}  )}) \\
& (- a_{2} \cos{ (\theta_{3s3}  )} - a_{3})  ) \\
\theta_{23s2}& = \operatorname{atan2} (Pz (- a_{2} \cos{ (\theta_{3s2}  )} - a_{3}) - \\
& (- Px \cos{ (\theta_{1s1}  )} - \\
& Py \sin{ (\theta_{1s1}  )}) (a_{2} \sin{ (\theta_{3s2}  )} - d_{4}),\\ 
& Pz (a_{2} \sin{ (\theta_{3s2}  )} - d_{4}) + \\
& (- Px \cos{ (\theta_{1s1}  )} - Py \sin{ (\theta_{1s1}  )}) \\
& (- a_{2} \cos{ (\theta_{3s2}  )} - a_{3})  ) \\
\theta_{23s3}& = \operatorname{atan2} (Pz (- a_{2} \cos{ (\theta_{3s1}  )} - a_{3}) - \\
& (- Px \cos{ (\theta_{1s1}  )} - \\
& Py \sin{ (\theta_{1s1}  )}) (a_{2} \sin{ (\theta_{3s1}  )} - d_{4}), \\
& Pz (a_{2} \sin{ (\theta_{3s1}  )} - d_{4}) + \\
& (- Px \cos{ (\theta_{1s1}  )} - Py \sin{ (\theta_{1s1}  )}) \\
& (- a_{2} \cos{ (\theta_{3s1}  )} - a_{3})  ) \\
\theta_{23s4}& = \operatorname{atan2} (Pz (- a_{2} \cos{ (\theta_{3s4}  )} - a_{3}) - \\
& (- Px \cos{ (\theta_{1s2}  )} - \\
& Py \sin{ (\theta_{1s2}  )}) (a_{2} \sin{ (\theta_{3s4}  )} - d_{4}),\\
& Pz (a_{2} \sin{ (\theta_{3s4}  )} - d_{4}) + \\
& (- Px \cos{ (\theta_{1s2}  )} - Py \sin{ (\theta_{1s2}  )}) \\
& (- a_{2} \cos{ (\theta_{3s4}  )} - a_{3})  )  
\end{align*}

\item $\theta_2$, chosen solver: algebraic solver

\begin{align*}
\theta_{2s1} &= \theta_{23s2} - \theta_{3s2} \\
\theta_{2s2} &= \theta_{23s4} - \theta_{3s4} \\
\theta_{2s3} &= \theta_{23s1} - \theta_{3s3} \\
\theta_{2s4} &= \theta_{23s3} - \theta_{3s1} 
\end{align*}

\item $\theta_4$, chosen solver: tangent
\begin{align*}
\theta_{4s1} &= \operatorname{atan2} (r_{13} \sin{ (\theta_{1s1}  )} - r_{23} \cos{ (\theta_{1s1}  )}, \\
& r_{13} \cos{ (\theta_{1s1}  )} \cos{ (\theta_{23s2}  )} + \\
& r_{23} \sin{ (\theta_{1s1}  )} \cos{ (\theta_{23s2}  )} - r_{33} \sin{ (\theta_{23s2}  )}  ) \\
\theta_{4s2} &= \operatorname{atan2} (- r_{13} \sin{ (\theta_{1s1}  )} + r_{23} \cos{ (\theta_{1s1}  )}, \\
& - r_{13} \cos{ (\theta_{1s1}  )} \cos{ (\theta_{23s2}  )} - \\
& r_{23} \sin{ (\theta_{1s1}  )} \cos{ (\theta_{23s2}  )} + r_{33} \sin{ (\theta_{23s2}  )  )} \\
\theta_{4s3} &= \operatorname{atan2} (r_{13} \sin{ (\theta_{1s2}  )} - r_{23} \cos{ (\theta_{1s2}  )}, \\
& r_{13} \cos{ (\theta_{1s2}  )} \cos{ (\theta_{23s4}  )} + \\
& r_{23} \sin{ (\theta_{1s2}  )} \cos{ (\theta_{23s4}  )} - r_{33} \sin{ (\theta_{23s4}  )}  ) \\
\theta_{4s4} &= \operatorname{atan2} (- r_{13} \sin{ (\theta_{1s2}  )} + r_{23} \cos{ (\theta_{1s2}  )}, \\
& - r_{13} \cos{ (\theta_{1s2}  )} \cos{ (\theta_{23s4}  )} - \\
& r_{23} \sin{ (\theta_{1s2}  )} \cos{ (\theta_{23s4}  )} + r_{33} \sin{ (\theta_{23s4}  )}  ) \\
\theta_{4s5} &= \operatorname{atan2} (r_{13} \sin{ (\theta_{1s2}  )} - r_{23} \cos{ (\theta_{1s2}  )}, \\
& r_{13} \cos{ (\theta_{1s2}  )} \cos{ (\theta_{23s1}  )} + \\
& r_{23} \sin{ (\theta_{1s2}  )} \cos{ (\theta_{23s1}  )} - r_{33} \sin{ (\theta_{23s1}  )}  ) \\
\theta_{4s6} &= \operatorname{atan2} (- r_{13} \sin{ (\theta_{1s2}  )} + r_{23} \cos{ (\theta_{1s2}  )}, \\
& - r_{13} \cos{ (\theta_{1s2}  )} \cos{ (\theta_{23s1}  )} -  \\
& r_{23} \sin{ (\theta_{1s2}  )} \cos{ (\theta_{23s1}  )} + r_{33} \sin{ (\theta_{23s1}  )}  ) \\
\theta_{4s7} &= \operatorname{atan2} (r_{13} \sin{ (\theta_{1s1}  )} - r_{23} \cos{ (\theta_{1s1}  )}, \\
& r_{13} \cos{ (\theta_{1s1}  )} \cos{ (\theta_{23s3}  )} + \\
& r_{23} \sin{ (\theta_{1s1}  )} \cos{ (\theta_{23s3}  )} - r_{33} \sin{ (\theta_{23s3}  )}  ) \\
\theta_{4s8} &= \operatorname{atan2} (- r_{13} \sin{ (\theta_{1s1}  )} + r_{23} \cos{ (\theta_{1s1}  )}, \\
& - r_{13} \cos{ (\theta_{1s1}  )} \cos{ (\theta_{23s3}  )} - \\
& r_{23} \sin{ (\theta_{1s1}  )} \cos{ (\theta_{23s3}  )} + r_{33} \sin{ (\theta_{23s3}  )}  ) 
\end{align*}

\item $\theta_5$, chosen solver: tangent
\begin{align*}
\theta_{5s1} &= \operatorname{atan2} (\frac{1}{\sin{ (\theta_{4s3}  )}} (- r_{13} \sin{ (\theta_{1s2}  )} + \\
& r_{23} \cos{ (\theta_{1s2}  )}), - r_{13} \sin{ (\theta_{23s4}  )} \cos{ (\theta_{1s2}  )} - \\
& r_{23} \sin{ (\theta_{1s2}  )} \sin{ (\theta_{23s4}  )} - r_{33} \cos{ (\theta_{23s4}  )}   \\
\theta_{5s2} &= \operatorname{atan2} (\frac{1}{\sin{ (\theta_{4s7}  )}} (- r_{13} \sin{ (\theta_{1s1}  )} + \\
& r_{23} \cos{ (\theta_{1s1}  )}), - r_{13} \sin{ (\theta_{23s3}  )} \cos{ (\theta_{1s1}  )} - \\
& r_{23} \sin{ (\theta_{1s1}  )} \sin{ (\theta_{23s3}  )} - r_{33} \cos{ (\theta_{23s3}  )}  ) \\
\theta_{5s3} &= \operatorname{atan2} (\frac{1}{\sin{ (\theta_{4s2}  )}} (- r_{13} \sin{ (\theta_{1s1}  )} + \\
& r_{23} \cos{ (\theta_{1s1}  )}), - r_{13} \sin{ (\theta_{23s2}  )} \cos{ (\theta_{1s1}  )} - \\
& r_{23} \sin{ (\theta_{1s1}  )} \sin{ (\theta_{23s2}  )} - r_{33} \cos{ (\theta_{23s2}  )}  ) \\
\theta_{5s4} &= \operatorname{atan2} (\frac{1}{\sin{ (\theta_{4s6}  )}} (- r_{13} \sin{ (\theta_{1s2}  )} + \\
& r_{23} \cos{ (\theta_{1s2}  )}),  - r_{13} \sin{ (\theta_{23s1}  )} \cos{ (\theta_{1s2}  )} - \\
& r_{23} \sin{ (\theta_{1s2}  )} \sin{ (\theta_{23s1}  )} - r_{33} \cos{ (\theta_{23s1}  )}  ) \\
\theta_{5s5} &= \operatorname{atan2} (\frac{1}{\sin{ (\theta_{4s4}  )}} (- r_{13} \sin{ (\theta_{1s2}  )} + \\
& r_{23} \cos{ (\theta_{1s2}  )}), - r_{13} \sin{ (\theta_{23s4}  )} \cos{ (\theta_{1s2}  )} - \\
& r_{23} \sin{ (\theta_{1s2}  )} \sin{ (\theta_{23s4}  )} - r_{33} \cos{ (\theta_{23s4}  )}  ) \\
\end{align*}
\begin{align*}
\theta_{5s6} &= \operatorname{atan2} (\frac{1}{\sin{ (\theta_{4s1}  )}} (- r_{13} \sin{ (\theta_{1s1}  )} + \\
& r_{23} \cos{ (\theta_{1s1}  )}), - r_{13} \sin{ (\theta_{23s2}  )} \cos{ (\theta_{1s1}  )} - \\
& r_{23} \sin{ (\theta_{1s1}  )} \sin{ (\theta_{23s2}  )} - r_{33} \cos{ (\theta_{23s2}  )}  ) \\
\theta_{5s7} &= \operatorname{atan2} (\frac{1}{\sin{ (\theta_{4s8}  )}} (- r_{13} \sin{ (\theta_{1s1}  )} + \\
& r_{23} \cos{ (\theta_{1s1}  )}), - r_{13} \sin{ (\theta_{23s3}  )} \cos{ (\theta_{1s1}  )} - \\
& r_{23} \sin{ (\theta_{1s1}  )} \sin{ (\theta_{23s3}  )} - r_{33} \cos{ (\theta_{23s3}  )}  ) \\
\theta_{5s8} &= \operatorname{atan2} (\frac{1}{\sin{ (\theta_{4s5}  )}} (- r_{13} \sin{ (\theta_{1s2}  )} + \\
& r_{23} \cos{ (\theta_{1s2}  )}),  - r_{13} \sin{ (\theta_{23s1}  )} \cos{ (\theta_{1s2}  )} - \\
& r_{23} \sin{ (\theta_{1s2}  )} \sin{ (\theta_{23s1}  )} - r_{33} \cos{ (\theta_{23s1}  )}  ) \\
\end{align*}

\item $\theta_6$, chosen solver: tangent
\begin{align*}
\theta_{6s1} &= \operatorname{atan2} (- \frac{1}{\sin{ (\theta_{5s4}  )}} (- r_{12} \sin{ (\theta_{23s1}  )} \cos{ (\theta_{1s2}  )} - \\
& r_{22} \sin{ (\theta_{1s2}  )} \sin{ (\theta_{23s1}  )} - r_{32} \cos{ (\theta_{23s1}  )}), \\
& \frac{1}{\sin{ (\theta_{5s4}  )}} (- r_{11} \sin{ (\theta_{23s1}  )} \cos{ (\theta_{1s2}  )} - \\
& r_{21} \sin{ (\theta_{1s2}  )} \sin{ (\theta_{23s1}  )} - r_{31} \cos{ (\theta_{23s1}  )})  ) \\
\theta_{6s2} &= \operatorname{atan2} (- \frac{1}{\sin{ (\theta_{5s8}  )}} (- r_{12} \sin{ (\theta_{23s1}  )} \cos{ (\theta_{1s2}  )} - \\
& r_{22} \sin{ (\theta_{1s2}  )} \sin{ (\theta_{23s1}  )} - r_{32} \cos{ (\theta_{23s1}  )}), \\
& \frac{1}{\sin{ (\theta_{5s8}  )}} (- r_{11} \sin{ (\theta_{23s1}  )} \cos{ (\theta_{1s2}  )} - \\
& r_{21} \sin{ (\theta_{1s2}  )} \sin{ (\theta_{23s1}  )} - r_{31} \cos{ (\theta_{23s1}  )})  ) \\
\theta_{6s3} &= \operatorname{atan2} (- \frac{1}{\sin{ (\theta_{5s1}  )}} (- r_{12} \sin{ (\theta_{23s4}  )} \cos{ (\theta_{1s2}  )} - \\
& r_{22} \sin{ (\theta_{1s2}  )} \sin{ (\theta_{23s4}  )} - r_{32} \cos{ (\theta_{23s4}  )}), \\
& \frac{1}{\sin{ (\theta_{5s1}  )}} (- r_{11} \sin{ (\theta_{23s4}  )} \cos{ (\theta_{1s2}  )} - \\
& r_{21} \sin{ (\theta_{1s2}  )} \sin{ (\theta_{23s4}  )} - r_{31} \cos{ (\theta_{23s4}  )})  ) \\
\theta_{6s4} &= \operatorname{atan2} (- \frac{1}{\sin{ (\theta_{5s2}  )}} (- r_{12} \sin{ (\theta_{23s3}  )} \cos{ (\theta_{1s1}  )} - \\
& r_{22} \sin{ (\theta_{1s1}  )} \sin{ (\theta_{23s3}  )} - r_{32} \cos{ (\theta_{23s3}  )}), \\
& \frac{1}{\sin{ (\theta_{5s2}  )}} (- r_{11} \sin{ (\theta_{23s3}  )} \cos{ (\theta_{1s1}  )} - \\
& r_{21} \sin{ (\theta_{1s1}  )} \sin{ (\theta_{23s3}  )} - r_{31} \cos{ (\theta_{23s3}  )})  ) \\
\theta_{6s5} &= \operatorname{atan2} (- \frac{1}{\sin{ (\theta_{5s6}  )}} (- r_{12} \sin{ (\theta_{23s2}  )} \cos{ (\theta_{1s1}  )} - \\
& r_{22} \sin{ (\theta_{1s1}  )} \sin{ (\theta_{23s2}  )} - r_{32} \cos{ (\theta_{23s2}  )}), \\
& \frac{1}{\sin{ (\theta_{5s6}  )}} (- r_{11} \sin{ (\theta_{23s2}  )} \cos{ (\theta_{1s1}  )} - \\
& r_{21} \sin{ (\theta_{1s1}  )} \sin{ (\theta_{23s2}  )} - r_{31} \cos{ (\theta_{23s2}  )})  ) \\
\end{align*}
\begin{align*}
\theta_{6s6} &= \operatorname{atan2} (- \frac{1}{\sin{ (\theta_{5s3}  )}} (- r_{12} \sin{ (\theta_{23s2}  )} \cos{ (\theta_{1s1}  )} - \\
& r_{22} \sin{ (\theta_{1s1}  )} \sin{ (\theta_{23s2}  )} - r_{32} \cos{ (\theta_{23s2}  )}), \\
& \frac{1}{\sin{ (\theta_{5s3}  )}} (- r_{11} \sin{ (\theta_{23s2}  )} \cos{ (\theta_{1s1}  )} - \\
& r_{21} \sin{ (\theta_{1s1}  )} \sin{ (\theta_{23s2}  )} - r_{31} \cos{ (\theta_{23s2}  )})  ) \\
\theta_{6s7} &= \operatorname{atan2} (- \frac{1}{\sin{ (\theta_{5s7}  )}} (- r_{12} \sin{ (\theta_{23s3}  )} \cos{ (\theta_{1s1}  )} - \\
& r_{22} \sin{ (\theta_{1s1}  )} \sin{ (\theta_{23s3}  )} - r_{32} \cos{ (\theta_{23s3}  )}), \\
& \frac{1}{\sin{ (\theta_{5s7}  )}} (- r_{11} \sin{ (\theta_{23s3}  )} \cos{ (\theta_{1s1}  )} - \\
& r_{21} \sin{ (\theta_{1s1}  )} \sin{ (\theta_{23s3}  )} - r_{31} \cos{ (\theta_{23s3}  )})  ) \\
\theta_{6s8} &= \operatorname{atan2} (- \frac{1}{\sin{ (\theta_{5s5}  )}} (- r_{12} \sin{ (\theta_{23s4}  )} \cos{ (\theta_{1s2}  )} - \\
&  r_{22} \sin{ (\theta_{1s2}  )} \sin{ (\theta_{23s4}  )} - r_{32} \cos{ (\theta_{23s4}  )}),\\
& \frac{1}{\sin{ (\theta_{5s5}  )}} (- r_{11} \sin{ (\theta_{23s4}  )} \cos{ (\theta_{1s2}  )} - \\
& r_{21} \sin{ (\theta_{1s2}  )} \sin{ (\theta_{23s4}  )} - r_{31} \cos{ (\theta_{23s4}  )})  ) \\
\end{align*}
\end{enumerate}
IKBT can find all 8 positions of PUMA 560, and the solution graph shows the dependency among variables in Fig. \ref{fig-Puma-sol}. And these joint poses can be grouped into sets:
\begin{align*}
Pose\quad 1 :& [ \theta_{1s1},  \theta_{2s4}, \theta_{3s1}, \theta_{4s8}, \theta_{5s7}, \theta_{6s7} ] \\
Pose\quad 2 :& [ \theta_{1s2},   \theta_{2s2}, \theta_{3s4},\theta_{4s4}, \theta_{5s5}, \theta_{6s8} ] \\
Pose\quad 3 :& [ \theta_{1s1},  \theta_{2s1}, \theta_{3s2}, \theta_{4s2}, \theta_{5s3}, \theta_{6s6} ] \\
Pose\quad 4 :& [ \theta_{1s2},   \theta_{2s3}, \theta_{3s3},\theta_{4s6}, \theta_{5s4}, \theta_{6s1} ] \\
Pose\quad 5 :& [ \theta_{1s2},  \theta_{2s3}, \theta_{3s3},\theta_{4s5}, \theta_{5s8}, \theta_{6s2} ] \\
Pose\quad 6 :& [ \theta_{1s1},   \theta_{2s1}, \theta_{3s2}, \theta_{4s1}, \theta_{5s6}, \theta_{6s5} ] \\
Pose\quad 7 :& [ \theta_{1s1},  \theta_{2s4}, \theta_{3s1}, \theta_{4s7}, \theta_{5s2}, \theta_{6s4} ] \\
Pose\quad 8 :& [ \theta_{1s2},  \theta_{2s2}, \theta_{3s4},\theta_{4s3}, \theta_{5s1}, \theta_{6s3} ]
\end{align*}

\subsubsection{Result Verification}
To verify the solution, we followed the process stated in section \ref{solution-verify}. The starting pose used is:
\[
\theta_1 = 30^\circ, \theta_2 = 50^\circ, \theta_3 = 40^\circ,
\]
\[
 \theta_4 = 45^\circ, \theta_5 = 120^\circ, \theta_6 = 60^\circ
\]

as well as the parameters:
\[
a_2 = 5, a_3 = 1, d_3 = 2, d_4 = 4
\]

We got numerical T matrix:
\[
T_d = 
\begin{bmatrix}
    -0.15720 & 0.97938 & 0.12682 & -1.68074 \\
    -0.59374 & -0.19635 & 0.78032 & 1.33902 \\
    0.78914 & 0.04737 & 0.61237 & -4.83022 \\
    0       & 0		 & 0		 &1 
\end{bmatrix}
\]

We then plug these numbers into symbolic solutions obtained from last section, and get poses listed in Table \ref{puma-num-pose}. Specifically, Pose 7 is considered the same as initial input pose, with differences $<10^{-4}$.  

With the numerical poses, we computed forward kinematics for each pose. T matrix computed from Pose 1 is selected as example, as it  shows the largest variation compared to the original T matrix: 

\[
\begin{bmatrix}
  -0.15720&    0.97939&    0.12682&   -1.68075\\
  -0.59374&   -0.19634&    0.78033&    1.33902\\
   0.78915&    0.04737&    0.61237&   -4.83025\\
   0&    0&    0&    1\\
\end{bmatrix}
\]

We got the same values compared to the original T matrix for all solution poses, with differences $\approx 10^{-5}$. This result unequivocally proves that IKBT's symbolic inverse kinematics analysis is correct.

\begin{table*}[t]
\centering
\begin{tabular}{ |c||c|c|c|c|c|c|  }
 \hline
 \multicolumn{7}{|c|}{Solution Poses} \\
 \hline
 Pose& $\theta_1$ & $\theta_2$ & $\theta_3$ & $\theta_4$ & $\theta_5$ & $\theta_6$ \\
 \hline
1 &-287.08771&  130.00008& -191.92745&   -6.78054&  -66.24462&    4.13687\\
2 &  29.99995&   49.99992&   39.99994& -135.00007& -119.99981& -120.00010\\
3 &  29.99995&  148.48625& -191.92745&  142.01103&   95.78709& -151.06756\\
4 &-287.08771&   31.51375&   39.99994&   18.00641&  159.53838&   18.33206\\
5 &-287.08771&  130.00008& -191.92745&  173.21946&   66.24462& -175.86313\\
6 &  29.99995&  148.48625& -191.92745&  -37.98897&  -95.78709&   28.93244\\
7 &  29.99995&   49.99992&   39.99994&   44.99993&  119.99981&   59.99990\\
8 &-287.08771&   31.51375&   39.99994& -161.99359& -159.53838& -161.66794\\
 \hline
\end{tabular}
\caption{\textbf{Puma 560 Numerical Solutions} }
\label{puma-num-pose}  
\end{table*}

\subsection{Example Solution - Robot without 3 intersecting axes}

\begin{table}[h]
\caption{Chair Helper DH parameters}\label{ChairHelperDH}
\begin{center}
\begin{tabular}{|c|c|c|c|c|}
\hline
Link & $\alpha_{N-1}$ & $a_{N-1}$ & $d_N$ & $\theta_N$ \\
\hline
1 &0 & 0 & $d_{1}$ & 0\\
\hline
2 & 0 & $l_{1}$ & 0 & $\theta_{2}$\\
\hline
3 & $\pi/2$ & 0 & $l_{2}$ & $\theta_3$\\
\hline
4 & $\pi/2$ & 0 & 0 & $\theta_4$\\
\hline
5 & - $\pi/2$ & 0 & $l_4$ & $\theta_5$\\
\hline
\end{tabular}
\end{center}
\end{table}

Previous software packages  which  perform inverse kinematics analysis usually require the robot to have three intersecting axes (such as the popular ROS package). To demonstrate IKBT's flexibility in handling robots with different configurations. We select the example of ``Chair Helper", a 5 DOF robot without three intersecting axes (Table \ref{ChairHelperDH})\footnote{Thanks to Prof. Melanie Shoemaker Plett.}.

Forward kinematics: 

\[
\begin{bmatrix}r_{11} & r_{12} & r_{13} & Px\\r_{21} & r_{22} & r_{23} & Py\\r_{31} & r_{32} & r_{33} & Pz\\0 & 0 & 0 & 1\end{bmatrix}
=  
\begin{bmatrix}
v_1 & v_2 & v_3 & v_4
\end{bmatrix}
\]

\[
v_1 =
\begin{bmatrix}
- c_{2} s_{3} s_{5} + c_{5} (c_{2} c_{3} c_{4} + s_{2} s_{4})\\
c_{5} (- c_{2} s_{4} + c_{3} c_{4} s_{2}) - s_{2} s_{3} s_{5} \\
c_{3} s_{5} + c_{4} c_{5} s_{3} \\
0
\end{bmatrix}
\]
\[
v_2 = 
\begin{bmatrix}
- c_{2} c_{5} s_{3} - s_{5} (c_{2} c_{3} c_{4} + s_{2} s_{4})\\
- c_{5} s_{2} s_{3} - s_{5} (- c_{2} s_{4} + c_{3} c_{4} s_{2}) \\
c_{3} c_{5} - c_{4} s_{3} s_{5}\\
0
\end{bmatrix}
\]
\[
v_3 = 
\begin{bmatrix}
- c_{2} c_{3} s_{4} + c_{4} s_{2}\\
- c_{2} c_{4} - c_{3} s_{2} s_{4} \\
- s_{3} s_{4} \\
0
\end{bmatrix}
\]

\[
v_4 =
\begin{bmatrix}
l_{1} + l_{2} s_{2} + l_{4} (- c_{2} c_{3} s_{4} + c_{4} s_{2})\\
- c_{2} c_{4} l_{4} - c_{2} l_{2} - c_{3} l_{4} s_{2} s_{4}\\
d_{1} - l_{4} s_{3} s_{4} \\
1
\end{bmatrix}
\]

Inverse kinematics solutions:

\begin{enumerate}
\item $d_1$, chosen solver: algebra
\begin{align*}
d_{1} &= Pz - l_{4} r_{33} 
\end{align*}

\item $\theta_{2}$, chosen solver: sine or cosine
\begin{align*}
\theta_{2s1} &= \operatorname{asin}{\left (\frac{1}{l_{2}} \left(Px - l_{1} - l_{4} r_{13}\right) \right )} \\
\theta_{2s2} &= - \operatorname{asin}{\left (\frac{1}{l_{2}} \left(Px - l_{1} - l_{4} r_{13}\right) \right )} + \pi 
\end{align*}

\item $\theta_{3}$, chosen solver: tangent
\begin{align*}
\theta_{3s1} &= \operatorname{atan2}{ (r_{33},r_{13} \cos{ (\theta_{2s2}  )} + r_{23} \sin{ (\theta_{2s2}  )}  )} \\
\theta_{3s2} &= \operatorname{atan2}{ (- r_{33},- r_{13} \cos{ (\theta_{2s2}  )} - r_{23} \sin{ (\theta_{2s2}  )}  )} \\
\theta_{3s3} &= \operatorname{atan2}{ (r_{33},r_{13} \cos{ (\theta_{2s1}  )} + r_{23} \sin{ (\theta_{2s1}  )}  )}\\
\theta_{3s4} &= \operatorname{atan2}{ (- r_{33},- r_{13} \cos{ (\theta_{2s1}  )} - r_{23} \sin{ (\theta_{2s1}  )}  )} \\
\end{align*}

\item $\theta_{4}$, chosen solver: tangent
\begin{align*}
\theta_{4s1} =& \operatorname{atan2} (- \frac{r_{33}}{\sin{ (\theta_{3s3}  )}}, \\
& r_{13} \sin{ (\theta_{2s1}  )} - r_{23} \cos{ (\theta_{2s1}  )}  ) \\
\theta_{4s2} =& \operatorname{atan2} (- \frac{r_{33}}{\sin{ (\theta_{3s2}  )}}, \\
& r_{13} \sin{ (\theta_{2s2}  )} - r_{23} \cos{ (\theta_{2s2}  )}  )\\
\theta_{4s3} =& \operatorname{atan2} (- \frac{r_{33}}{\sin{ (\theta_{3s1}  )}}, \\
& r_{13} \sin{ (\theta_{2s2}  )} - r_{23} \cos{ (\theta_{2s2}  )}  ) \\
\theta_{4s4} =& \operatorname{atan2} (- \frac{r_{33}}{\sin{ (\theta_{3s4}  )}}, \\
& r_{13} \sin{ (\theta_{2s1}  )} - r_{23} \cos{ (\theta_{2s1}  )}  ) \\
\end{align*}

\item $\theta_{5}$, chosen solver: tangent
\begin{align*}
\theta_{5s1} =& \operatorname{atan2} (\frac{1}{\sin{ (\theta_{4s3}  )}} (- r_{12} \sin{ (\theta_{2s2}  )} + \\
& r_{22} \cos{ (\theta_{2s2}  )}),\frac{1}{\sin{ (\theta_{4s3}  )}} (r_{11} \sin{ (\theta_{2s2}  )} - \\
& r_{21} \cos{ (\theta_{2s2}  )})  ) \\
\theta_{5s2} =& \operatorname{atan2} (\frac{1}{\sin{ (\theta_{4s4}  )}} (- r_{12} \sin{ (\theta_{2s1}  )} + \\
& r_{22} \cos{ (\theta_{2s1}  )}),\frac{1}{\sin{ (\theta_{4s4}  )}} (r_{11} \sin{ (\theta_{2s1}  )} - \\
& r_{21} \cos{ (\theta_{2s1}  )})  ) \\
\theta_{5s3} =& \operatorname{atan2} (\frac{1}{\sin{ (\theta_{4s1}  )}} (- r_{12} \sin{ (\theta_{2s1}  )} + \\
& r_{22} \cos{ (\theta_{2s1}  )}),\frac{1}{\sin{ (\theta_{4s1}  )}} (r_{11} \sin{ (\theta_{2s1}  )} - \\
& r_{21} \cos{ (\theta_{2s1}  )})  ) \\
\theta_{5s4} =& \operatorname{atan2} (\frac{1}{\sin{ (\theta_{4s2}  )}} (- r_{12} \sin{ (\theta_{2s2}  )} + \\
&r_{22} \cos{ (\theta_{2s2}  )}),\frac{1}{\sin{ (\theta_{4s2}  )}} (r_{11} \sin{ (\theta_{2s2}  )} - \\
&r_{21} \cos{ (\theta_{2s2}  )})  ) \\
\end{align*}
\end{enumerate}

Solutions sets:
\begin{align*}
[d_{1}, \theta_{2s1}, \theta_{3s3}, \theta_{4s1}, \theta_{5s3}] \\
[d_{1}, \theta_{2s1}, \theta_{3s4}, \theta_{4s4}, \theta_{5s2}] \\
[d_{1}, \theta_{2s2}, \theta_{3s1}, \theta_{4s3}, \theta_{5s1}] \\
[d_{1}, \theta_{2s2}, \theta_{3s2}, \theta_{4s2}, \theta_{5s4}]
\end{align*}

Numerical verification confirmed that the inverse kinematics solutions are correct. 
\subsection{Example Solution - Robot with strictly solution graph}
\begin{figure*}[ht]
    \centering
    \includegraphics[width = 0.8\textwidth]{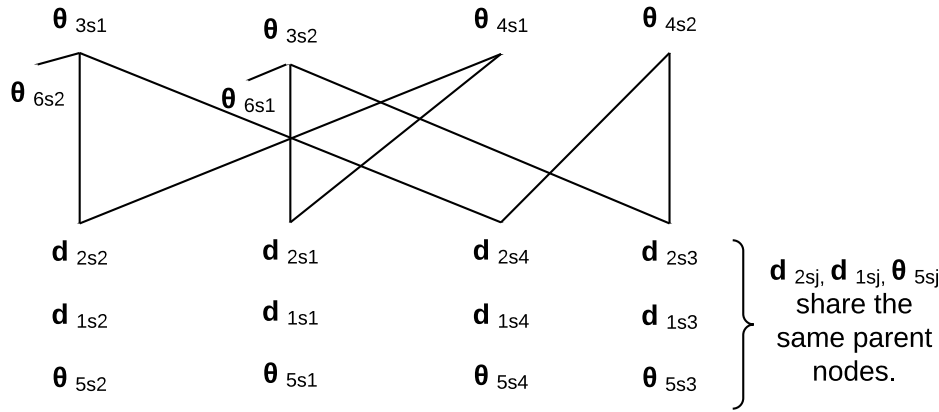}
    \caption{\textbf{Olson13 Solution Graph.}}
    \label{fig-olson13-sol}
\end{figure*}

The following example (Olson13) illustrates the necessity of a graph when tracking dependency, where variables have two independent parent variables, as shown in \ref{fig-olson13-sol}. DH parameters are listed in Table \ref{Olson13DH}. Olson13 is a 6-DOF robot. The unknown variables are: $[d_1 \quad d_2 \quad \theta_3 \quad \theta_4 \quad \theta_5 \quad \theta_6]$. The known parameters are: $[l_3 \quad l_4 \quad l_5]$

\begin{table}[h]
\caption{Olson13 DH parameters}\label{Olson13DH}
\begin{center}
\begin{tabular}{|c|c|c|c|c|}
\hline
Link & $\alpha_{N-1}$ & $a_{N-1}$ & $d_N$ & $\theta_N$ \\
\hline
1 & $- \pi/2$ & 0 & $d_{1}$ & $\pi/2$\\
\hline
2 & $\pi/2$ & 0 & $d_{2}$ & $- \pi/2$\\
\hline
3 & $\pi/2$ & 0 & $l_{3}$ & $\theta_{3}$\\
\hline
4 & $\pi/2$ & 0 & 0 & $\theta_{4}$\\
\hline
5 & 0 & $l_{4}$ & 0 & $\theta_{5}$\\
\hline
6 & $\pi/2$ & 0 & $l_{5}$ & $\theta_{6}$\\
\hline
\end{tabular}
\end{center}
\end{table}

Forward kinematics:

\[
\begin{bmatrix}r_{11} & r_{12} & r_{13} & Px\\r_{21} & r_{22} & r_{23} & Py\\r_{31} & r_{32} & r_{33} & Pz\\0 & 0 & 0 & 1\end{bmatrix} =  
\begin{bmatrix} v_1 & v_2 & v_3 & v_4\end{bmatrix}
\]

\[
\begin{bmatrix}
v_1 & v_2
\end{bmatrix}
=
\begin{bmatrix}
- c_{3} s_{6} + c_{45} c_{6} s_{3} & - c_{3} c_{6} - c_{45} s_{3} s_{6}\\
- c_{3} c_{45} c_{6} - s_{3} s_{6} & c_{3} c_{45} s_{6} - c_{6} s_{3} \\
c_{6} s_{45} & - s_{45} s_{6}\\
0 & 0
\end{bmatrix}
\]

\[
\begin{bmatrix}
v_3 & v_4
\end{bmatrix}
=
\begin{bmatrix}
 s_{3} s_{45} & c_{4} l_{4} s_{3} + d_{2} + l_{5} s_{3} s_{45}\\
- c_{3} s_{45} & - c_{3} c_{4} l_{4} - c_{3} l_{5} s_{45} + d_{1}\\
 - c_{45} & - c_{45} l_{5} + l_{3} + l_{4} s_{4}\\
0 & 1
\end{bmatrix}
\]

The variables are solved in the following order: 
\begin{enumerate}
\item $\theta_3$, chosen solver: tangent
\begin{align*}
\theta_{3s1} &= \operatorname{atan2}{\left (- r_{13},r_{23} \right )} \\
\theta_{3s2} &= \operatorname{atan2}{\left (r_{13},- r_{23} \right )} 
\end{align*}

\item $\theta_4$, chosen solver: sine or cosine
\begin{align*}
\theta_{4s1} &= \operatorname{asin}{\left (\frac{1}{l_{4}} \left(Pz - l_{3} - l_{5} r_{33}\right) \right )} \\
\theta_{4s2} &= - \operatorname{asin}{\left (\frac{1}{l_{4}} \left(Pz - l_{3} - l_{5} r_{33}\right) \right )} + \pi 
\end{align*}

\item $\theta_5$, chosen solver: tangent
\begin{align*}
\theta_{5s1} &= \operatorname{atan2} (r_{13} \sin{ (\theta_{3s2}  )} \cos{ (\theta_{4s1}  )} - \\
& r_{23} \cos{ (\theta_{3s2}  )} \cos{ (\theta_{4s1}  )} + r_{33} \sin{ (\theta_{4s1}  )}, \\
& r_{13} \sin{ (\theta_{3s2}  )} \sin{ (\theta_{4s1}  )} - \\
& r_{23} \sin{ (\theta_{4s1}  )} \cos{ (\theta_{3s2}  )} - r_{33} \cos{ (\theta_{4s1}  )}  ) \\
\theta_{5s2} &= \operatorname{atan2} (r_{13} \sin{ (\theta_{3s1}  )} \cos{ (\theta_{4s1}  )} - \\ & r_{23} \cos{ (\theta_{3s1}  )} \cos{ (\theta_{4s1}  )} + r_{33} \sin{ (\theta_{4s1}  )}, \\
& r_{13} \sin{ (\theta_{3s1}  )} \sin{ (\theta_{4s1}  )} - \\
& r_{23} \sin{ (\theta_{4s1}  )} \cos{ (\theta_{3s1}  )} - r_{33} \cos{ (\theta_{4s1}  )}  ) \\
\theta_{5s3} &= \operatorname{atan2} (r_{13} \sin{ (\theta_{3s2}  )} \cos{ (\theta_{4s2}  )} - \\ & r_{23} \cos{ (\theta_{3s2}  )} \cos{ (\theta_{4s2}  )} + r_{33} \sin{ (\theta_{4s2}  )}, \\
& r_{13} \sin{ (\theta_{3s2}  )} \sin{ (\theta_{4s2}  )} - \\
& r_{23} \sin{ (\theta_{4s2}  )} \cos{ (\theta_{3s2}  )} - r_{33} \cos{ (\theta_{4s2}  )}  ) \\
\theta_{5s4} &= \operatorname{atan2} (r_{13} \sin{ (\theta_{3s1}  )} \cos{ (\theta_{4s2}  )} -\\
& r_{23} \cos{ (\theta_{3s1}  )} \cos{ (\theta_{4s2}  )} + r_{33} \sin{ (\theta_{4s2}  )},\\
& r_{13} \sin{ (\theta_{3s1}  )} \sin{ (\theta_{4s2}  )} - \\
& r_{23} \sin{ (\theta_{4s2}  )} \cos{ (\theta_{3s1}  )} - r_{33} \cos{ (\theta_{4s2}  )}  ) \\
\end{align*}

\item $\theta_6$, chosen solver: tangent
\begin{align*}
\theta_{6s1} &= \operatorname{atan2} (- r_{11} \cos{ (\theta_{3s2}  )} - r_{21} \sin{ (\theta_{3s2}  )}, \\
& - r_{12} \cos{ (\theta_{3s2}  )} - r_{22} \sin{ (\theta_{3s2}  )}  ) \\
\theta_{6s2} &= \operatorname{atan2} (- r_{11} \cos{ (\theta_{3s1}  )} - r_{21} \sin{ (\theta_{3s1}  )}, \\
& - r_{12} \cos{ (\theta_{3s1}  )} - r_{22} \sin{ (\theta_{3s1}  )}  ) \\
\end{align*}
\item $d_1$, chosen solver: algebra
\begin{align*}
d_{1s1} &= Py + l_{4} \cos{\left (\theta_{3s2} \right )} \cos{\left (\theta_{4s1} \right )} - l_{5} r_{23} \\
d_{1s2} &= Py + l_{4} \cos{\left (\theta_{3s1} \right )} \cos{\left (\theta_{4s1} \right )} - l_{5} r_{23} \\
d_{1s3} &= Py + l_{4} \cos{\left (\theta_{3s2} \right )} \cos{\left (\theta_{4s2} \right )} - l_{5} r_{23} \\
d_{1s4} &= Py + l_{4} \cos{\left (\theta_{3s1} \right )} \cos{\left (\theta_{4s2} \right )} - l_{5} r_{23} \\
\end{align*}
\item $d_2$, chosen solver: algebra
\begin{align*}
d_{2s1} &= Px - l_{4} \sin{\left (\theta_{3s2} \right )} \cos{\left (\theta_{4s1} \right )} - l_{5} r_{13} \\
d_{2s2} &= Px - l_{4} \sin{\left (\theta_{3s1} \right )} \cos{\left (\theta_{4s1} \right )} - l_{5} r_{13} \\
d_{2s3} &= Px - l_{4} \sin{\left (\theta_{3s2} \right )} \cos{\left (\theta_{4s2} \right )} - l_{5} r_{13}\\ 
d_{2s4} &= Px - l_{4} \sin{\left (\theta_{3s1} \right )} \cos{\left (\theta_{4s2} \right )} - l_{5} r_{13} \\
\end{align*}
\end{enumerate}

The following are the sets of joint solutions (poses) for this manipulator:

\begin{align*}
[d_{1s3}, d_{2s3}, \theta_{3s2}, \theta_{4s2}, \theta_{5s3}, \theta_{6s1}]\\
[d_{1s4}, d_{2s4}, \theta_{3s1}, \theta_{4s2}, \theta_{5s4}, \theta_{6s2}]\\
[d_{1s1}, d_{2s1}, \theta_{3s2}, \theta_{4s1}, \theta_{5s1}, \theta_{6s1}]\\
[d_{1s2}, d_{2s2}, \theta_{3s1}, \theta_{4s1}, \theta_{5s2}, \theta_{6s2}]
\end{align*}

The solution graph is shown in Fig. \ref{fig-olson13-sol}. $d_1$, $d_2$, and $\theta_5$ all share two independently solved parent variables: $\theta_3$ and $\theta_4$. Thus, it results in a dependency graph.

\section{Discussion}
\subsection{Related Work and Comparison}
Previous research on inverse kinematics lays a good foundation for IKBT.  \cite{Herrera-Bendezu} used a rule-based pattern matching approach (implemented as an expert system in LISP) from which IKBT adapted the sin or cos solver, tangent solver, and simultaneous equation solver.  Similar to IKBT, that system scanned a list of equations, and found the ones matching  patterns, then fetching the respective solutions. Their method solved several commercial robots, including the more complicated PUMA 560.  Limitations of this work include a hard coded framework for solution sequencing, and dependence on obsolescent software. Comparatively, IKBT uses very few rule-based solvers, indicative of more efficient logical reasoning.  

\cite{Wenz} used a similar approach to ours. Their system also uses rule-based solvers implemented in LISP, though the detailed rules or the source code were not made public. 
The system of \cite{Wenz} solves equations sequentially and stops working on a variable as soon as it is solved. 
In contrast, IKBT's assigner node picks a variable first, and tries the entire toolbox for the chosen variable. We designed IKBT this way to get the optimal solution, in the  situation where more than one solver applies to the same variable (choosing different equations). IKBT ranks the solutions obtained and chooses the best solution. \cite{Herrera-Bendezu,Wenz} did not show capability of finding all possible solution or tracking dependency among variables.

\cite{Halperin} used a different approach of converting the set of kinematics equations into a univariate polynomial using elimination techniques. It is effective in solving specific robots. However, whether their methods could be applied to other robots was not systematically tested. Also, because it uses an approach unlike what human experts do, it is harder to check the correctness of the solution or strategy: IKBT's toolbox contains only well known rules  frequently used by human experts.   

\cite{IMChen}'s solver used a product-of-exponentials formula, which doesn't require D-H parameters, and is robust in dealing with kinematics singularities. However, it only showed the capability of handling numerical inputs and rendering numerical solutions, and very limited solving capability for complex robots (6-DOF robots $\approx$ 50\%) . Compared to \cite{IMChen}, IKBT handles symbolic input, generates closed-form solutions, and achieved much better success rate with complex 5-DOF (100 \%) and 6-DOF (80\%) robots. \cite{Chapelle}  used evolutionary algorithms to get an approximate closed-form IK solution.  By contrast, IKBT computes exact symbolic closed-form solutions. 

IKFast performs symbolic inverse kinematics analysis as part of the OpenRAVE package \cite{Diankov}. Instead of general solving techniques, IKFast adapts a case-specific approach. It categories robots by their number of DOFs, and uses DOF-specific hard-coded algorithms for arms with different DOFs.  IKFast generates a ``dependency tree".  However, a tree cannot represent the multiple
independent dependencies we found for some variables in some robots. The graph-based representation generated by IKBT solves this problem.  We have tried to run IKFast to compare the performance differences with IKBT, but we encountered several problems: First, it requires the full installation of the OpenRave suite. Second, its exclusive dependency on older version of software packages is incompatible with current versions. In order to run IKFast, one needs to downgrade Python and related packages.  

One distinguishing feature of IKBT is the generation of solution graph of joint variables, though in some above-mentioned studies "dependency graph" was explored in other unrelated context. As described in details in section Solution Graph,  dependency graph is essential to keep track of the joint poses, and make sure the solutions sets are complete and necessary. Complete means IKBT can find all possible joint poses if the solution exists. Necessary means that all solutions are unique, not duplicate to each other. Dependency tracking was usually done by engineers, to the best of our knowledge, IKBT is the first to automate this process.   

\subsection{Perspectives}

Building on top of the previous research, IKBT has several defining advantages including applicability to any robot (up to 6-DOF), generalized solving scheme,  extensible toolbox, modern and easy to implement language (Python), and dependencies limited to only a few libraries. We expect these characteristics will spur the wide adoption of IKBT into the robotics research and education communities.
 
Rule-based solvers included in IKBT's toolbox are commonly employed by human experts when solving inverse kinematics problems. This is advantageous because IKBT is not limited by robot configuration, specifically, it doesn't require three orthogonal axes in order to solve a robot. 

IKBT's Behavior Tree represents an interpretable strategy - vital for judging many AI applications. This makes it easier to examine the correctness of the solution and the strategy formulating process. Although IKBT's approach costs more computing time than DOF-specific algorithms (4 ms, according to \cite{Diankov}, symbolic derivation only has to be done once per robot arm design.  The Behavior Tree is easily modified and the solver toolbox is readily extensible.  Although all the results presented here were generated by the BT of Fig. \ref{fig-BTStructure}, it may be the case that a custom Behavior Tree could solve additional robots or solve robots more efficiently. 

Behavior Tree has gained great success in game AI \cite{BTPlatformGame,emotion-BT}, and showed substantial possibilities in robotics research \cite{OgrenUAV,marzinotto2014towards,ISR16Colledanchise,DYHu-BT}. IKBT serves as a proof-of-concept of solving high-cognitive problems with Behavior Tree. IKBT mimics human experts' logical reasoning process, and constructs a generalized solving scheme applicable to an entire class of problems, using a small number of knowledge leafs. While most of current AI work focuses on recognizing and understanding scenarios, Behavior Tree emerges as a path to strengthen an equally vital component - logical reasoning. In IKBT, all knowledge-based solvers are coded by us, in other words, we "teach" the system about all the pre-existing rules and tricks people use when solving inverse kinematics problems. Given the current state of AI development, it is feasible to let the system to learn the knowledge by itself, through observing patterns and understanding the meaning behind. Combining forces of recognition, learning, and reasoning, we might be on our way to unlock the next level of autonomy.


%

\appendices

\ifCLASSOPTIONcompsoc
  \section*{Acknowledgments}
\else
  \section*{Acknowledgment}
\fi

We gratefully acknowledge support from National Science Foundation grant 1637444 and support for Blake Hannaford at Google-X / Google Life-Sciences / Verily in 2015.

\ifCLASSOPTIONcaptionsoff
  \newpage
\fi



\bibliographystyle{IEEEtran}
\bibliography{IKBTbib}

\begin{thebibliography}{10}
\providecommand{\url}[1]{#1}
\csname url@samestyle\endcsname
\providecommand{\newblock}{\relax}
\providecommand{\bibinfo}[2]{#2}
\providecommand{\BIBentrySTDinterwordspacing}{\spaceskip=0pt\relax}
\providecommand{\BIBentryALTinterwordstretchfactor}{4}
\providecommand{\BIBentryALTinterwordspacing}{\spaceskip=\fontdimen2\font plus
\BIBentryALTinterwordstretchfactor\fontdimen3\font minus
  \fontdimen4\font\relax}
\providecommand{\BIBforeignlanguage}[2]{{%
\expandafter\ifx\csname l@#1\endcsname\relax
\typeout{** WARNING: IEEEtran.bst: No hyphenation pattern has been}%
\typeout{** loaded for the language `#1'. Using the pattern for}%
\typeout{** the default language instead.}%
\else
\language=\csname l@#1\endcsname
\fi
#2}}
\providecommand{\BIBdecl}{\relax}
\BIBdecl

\bibitem{matlab}
P.~I. Corke, ``A robotics toolbox for matlab,'' \emph{IEEE Robotics Automation
  Magazine}, vol.~3, no.~1, pp. 24--32, Mar 1996.

\bibitem{ROB:ROB4620070406}
\BIBentryALTinterwordspacing
L.~Kelmar and P.~K. Khosla, ``Automatic generation of forward and inverse
  kinematics for a reconfigurable modular manipulator system,'' \emph{Journal
  of Robotic Systems}, vol.~7, no.~4, pp. 599--619, 1990. [Online]. Available:
  \url{http://dx.doi.org/10.1002/rob.4620070406}
\BIBentrySTDinterwordspacing

\bibitem{Herrera-Bendezu}
L.~G. Herrera-Bendezu, E.~Mu, and J.~T. Cain, ``Symbolic computation of robot
  manipulator kinematics,'' in \emph{Proceedings. 1988 IEEE International
  Conference on Robotics and Automation}, Apr 1988, pp. 993--998 vol.2.

\bibitem{Halperin}
D.~Halperin, ``Automatic kinematic modelling of robot manipulators and symbolic
  generation of their inverse kinematics solutions (extended abstract),'' pp.
  310--317, 1991.

\bibitem{lim2010evolving}
C.-U. Lim, R.~Baumgarten, and S.~Colton, ``Evolving behaviour trees for the
  commercial game defcon,'' in \emph{European Conference on the Applications of
  Evolutionary Computation}.\hskip 1em plus 0.5em minus 0.4em\relax Springer,
  2010, pp. 100--110.

\bibitem{marzinotto2014towards}
A.~Marzinotto, M.~Colledanchise, C.~Smith, and P.~Ogren, ``Towards a unified
  behavior trees framework for robot control,'' in \emph{Robotics and
  Automation (ICRA), 2014 IEEE International Conference on}.\hskip 1em plus
  0.5em minus 0.4em\relax IEEE, 2014, pp. 5420--5427.

\bibitem{IROS17Colledanchise}
M.~Colledanchise, M.~Richard, and P.~Ogren, ``{Synthesis of
  Correct-by-Construction Behavior Trees},'' in \emph{Intelligent {R}obots and
  {S}ystems ({IROS} 2017), 2017 {IEEE/RSJ} {I}nternational {C}onference on},
  Sept 2017, pp. 1482--1488.

\bibitem{ISR16Colledanchise}
M.~Colledanchise, A.~Marzinotto, D.~V. Dimarogonas, and P.~Ogren, ``The
  advantages of using behavior trees in multi-robot systems,'' in
  \emph{International {S}ymposium on {R}obotics ({ISR})}, June 2016.

\bibitem{colledanchise2014performance}
M.~Colledanchise, A.~Marzinotto, and P.~Ogren, ``Performance analysis of
  stochastic behavior trees,'' in \emph{2014 IEEE International Conference on
  Robotics and Automation (ICRA)}.\hskip 1em plus 0.5em minus 0.4em\relax IEEE,
  2014, pp. 3265--3272.

\bibitem{bagnell2012integrated}
J.~A. Bagnell, F.~Cavalcanti, L.~Cui, T.~Galluzzo, M.~Hebert, M.~Kazemi,
  M.~Klingensmith, J.~Libby, T.~Y. Liu, N.~Pollard, M.~Pivtoraiko, J.~S.
  Valois, and R.~Zhu, ``An integrated system for autonomous robotics
  manipulation,'' in \emph{2012 IEEE/RSJ International Conference on
  Intelligent Robots and Systems}.\hskip 1em plus 0.5em minus 0.4em\relax IEEE,
  2012, pp. 2955--2962.

\bibitem{guerin2015framework}
K.~R. Guerin, C.~Lea, C.~Paxton, and G.~D. Hager, ``A framework for end-user
  instruction of a robot assistant for manufacturing,'' in \emph{Robotics and
  Automation (ICRA), 2015 IEEE International Conference on}.\hskip 1em plus
  0.5em minus 0.4em\relax IEEE, 2015, pp. 6167--6174.

\bibitem{DYHu-BT}
D.~Hu, Y.~Gong, B.~Hannaford, and E.~J. Seibel, ``Semi-autonomous simulated
  brain tumor ablation with ravenii surgical robot using behavior tree,'' in
  \emph{2015 IEEE International Conference on Robotics and Automation (ICRA)},
  May 2015, pp. 3868--3875.

\bibitem{DBLP:journals/corr/HannafordHZL16}
\BIBentryALTinterwordspacing
B.~Hannaford, D.~Hu, D.~Zhang, and Y.~Li, ``Simulation results on selector
  adaptation in behavior trees,'' \emph{CoRR}, vol. abs/1606.09219, 2016.
  [Online]. Available: \url{http://arxiv.org/abs/1606.09219}
\BIBentrySTDinterwordspacing

\bibitem{Craig:1989:IRM:534661}
J.~J. Craig, \emph{Introduction to Robotics: Mechanics and Control},
  2nd~ed.\hskip 1em plus 0.5em minus 0.4em\relax Boston, MA, USA:
  Addison-Wesley Longman Publishing Co., Inc., 1989.

\bibitem{Wenz}
M.~Wenz and H.~Worn, ``Solving the inverse kinematics problem symbolically by
  means of knowledge-based and linear algebra-based methods,'' in \emph{2007
  IEEE Conference on Emerging Technologies and Factory Automation (EFTA 2007)},
  Sept 2007, pp. 1346--1353.

\bibitem{IMChen}
I.-M. Chen and Y.~Gao, ``Closed-form inverse kinematics solver for
  reconfigurable robots,'' in \emph{Proceedings 2001 ICRA. IEEE International
  Conference on Robotics and Automation (Cat. No.01CH37164)}, vol.~3, 2001, pp.
  2395--2400 vol.3.

\bibitem{Chapelle}
F.~Chapelle and P.~Bidaud, ``A closed form for inverse kinematics approximation
  of general 6r manipulators using genetic programming,'' in \emph{Proceedings
  2001 ICRA. IEEE International Conference on Robotics and Automation (Cat.
  No.01CH37164)}, vol.~4, 2001, pp. 3364--3369 vol.4.

\bibitem{Diankov}
R.~Diankov, ``Automated construction of robotic manipulation programs,'' 2010.

\bibitem{BTPlatformGame}
M.~Nicolau, D.~Perez-Liebana, M.~O’Neill, and A.~Brabazon, ``Evolutionary
  behavior tree approaches for navigating platform games,'' \emph{IEEE
  Transactions on Computational Intelligence and AI in Games}, vol.~9, no.~3,
  pp. 227--238, Sept 2017.

\bibitem{emotion-BT}
A.~Johansson and P.~Dell'Acqua, ``Emotional behavior trees,'' in \emph{2012
  IEEE Conference on Computational Intelligence and Games (CIG)}, Sept 2012,
  pp. 355--362.

\bibitem{OgrenUAV}
P.~Ogren, ``Increasing modularity of uav control systems using computer game
  behavior trees,'' 08 2012.

\end{thebibliography}

\end{document}